\documentclass[pdflatex,sn-mathphys-num]{sn-jnl}

\usepackage{graphicx}%
\usepackage{multirow}%
\usepackage{amsmath,amssymb,amsfonts}%
\usepackage{amsthm}%
\usepackage{mathrsfs}%
\usepackage[title]{appendix}%
\usepackage{xcolor}%
\usepackage{textcomp}%
\usepackage{manyfoot}%
\usepackage{booktabs}%
\usepackage{algorithm}%
\usepackage{algorithmicx}%
\usepackage{algpseudocode}%
\usepackage{listings}%
\usepackage[table]{xcolor}
\usepackage{booktabs}
\usepackage{amssymb}
\usepackage{booktabs}
\usepackage{tabularx}
\usepackage[table]{xcolor}
\usepackage{colortbl}
\usepackage{siunitx}
\usepackage{amssymb}   
\usepackage{xcolor}
\usepackage{booktabs}
\usepackage[table]{xcolor}

\definecolor{cCorrect}{HTML}{526A78}
\definecolor{cPartial}{HTML}{B8A99A}
\definecolor{cIncorrect}{HTML}{C47772}

\definecolor{stargold}{HTML}{F5B700}
\newcommand{\goldstar}{\textcolor{stargold}{$\bigstar$}}
\definecolor{navy}{HTML}{344E5C}
\definecolor{pale}{HTML}{EEF2F4}
\definecolor{accent}{HTML}{A66B55}

\theoremstyle{thmstyleone}%
\theoremstyle{thmstyletwo}%

\theoremstyle{thmstylethree}%

\usepackage[normalem]{ulem} 
\usepackage{hyperref} 
\usepackage{comment}

\begin{document}

\title[Article Title]{Hx2Dx: From Anamnesis to Diagnosis - A Multimodal Arena for Conversational Clinical Reasoning in LLMs}


\author*[1]{\fnm{Federico} \sur{Ravenda}}\email{federico.ravenda@usi.ch}

\author[2]{\fnm{Mirko } \sur{Cesarini}}

\author[1,3]{\fnm{Daniele} \sur{Montagnani}}

\author[1]{\fnm{Volodymyr} \sur{Karpenko}}

\author[1,4]{\fnm{Antonietta} \sur{Mira}}
\equalcont{Equal Supervision.}

\author[1,5]{\fnm{Andrea} \sur{Raballo}}
\equalcont{Equal Supervision.}

\affil[1]{\orgdiv{REMEDI Lab}, \orgname{Università della Svizzera italiana}, \city{Lugano}, \country{Switzerland}}

\affil[2]{\orgdiv{Department of Statistics and
Quantitative Methods}, \orgname{University of Milano-Bicocca}, \city{Milan}, \country{Italy}}

\affil[3]{\orgdiv{Department of Mathematics}, 
\orgname{Pavia University}, \orgaddress{\city{Pavia},  \country{Italy}}}

\affil[4]{\orgdiv{Department of Science and High Technology}, \orgname{Insubria University}, \orgaddress{\city{Como},  \country{Italy}}}

\affil[5]{\orgdiv{Cantonal Sociopsychiatric Organisation, Public Health Division, Department of Health and Social Care}, \orgname{Repubblica e Cantone Ticino}, \orgaddress{\city{Mendrisio},  \country{Switzerland}}}

\abstract{Large language models (LLMs) achieve expert-level diagnostic accuracy in static evaluations, where the model receives a curated vignette or case report that already condenses the diagnostically relevant findings. 
Such evaluations measure inference over already selected evidence; the evaluations do not account for the inquiry used to obtain it.
In clinical practice, however, diagnosis is an iterative process of inquiry. Clinicians take the patient’s history, or anamnesis (\textbf{Hx}), through targeted questioning, examine the patient, and request and interpret tests, progressively narrowing the differential until they reach a final diagnosis (\textbf{Dx}). We introduce Hx2Dx ARENA, an interactive benchmark of 1'000 clinical cases from the New England Journal of Medicine (NEJM). Each case is a simulated encounter that proceeds from initial presentation through history taking and physical examination to diagnostic workup. We evaluate open- and closed-source LLMs under multiple configurations. Performance varies considerably across models in the interactive setting, and every model loses diagnostic accuracy relative to the static setting, where the full case is presented upfront. The ability to ask the right questions accounts for much of both the gap between models and the loss of accuracy: diagnosis improves as questions become more targeted and build a more complete clinical picture. LLMs thus reason well once the relevant data are available but remain limited in hypothesis-driven information gathering, in knowing what to ask in order to know. Closing this gap is a prerequisite for their safe integration into real-world diagnostic workflows.}

\keywords{Large Language Models, Medical Multimodal Benchmark, Clinical Conversation, Diagnostic Accuracy}



\maketitle

\section{Introduction}

Diagnosis is a hypothetico-deductive process, not an act of pattern recognition. From the chief complaint, the physician forms early hypotheses and selects the questions, examination maneuvers, and investigations most likely to discriminate between them. 

Each finding revises the differential and directs the next step of the workup. Accuracy depends as much on knowing which findings to pursue as on interpreting them.
Current LLM evaluations test mostly the second half of this process. Models perform remarkably well on clinical benchmarks~\cite{singhal2023large,singhal2025toward}, at times matching or exceeding physicians~\cite{raballo2025diagnosing}, but they receive cases that have already been worked up: the history taken, pertinent positives and negatives selected, the right tests ordered. 

At the bedside, discriminating findings are rarely volunteered; they must be elicited. The gap widens in multimodal cases, where evidence spans history, examination, laboratory values, ECG, imaging and pathology, and what is obtained, and in which order, depends on the working hypothesis.

Interactive environments that place LLMs in simulated encounters~\cite{tu2025towards,li2024mediq} have shown that diagnostic performance depends on how the encounter is conducted, not only on medical knowledge. However, they have not quantified systematically, across a large and diverse case set, in a multimodal setting, how much accuracy is lost when a model must reconstruct the case through the encounter rather than receive it complete.

We introduce \textbf{Hx2Dx ARENA}, an interactive benchmark that evaluates LLMs as diagnosticians under partial observability. The arena comprises 1'000 clinical cases from the \textit{New England Journal of Medicine} (NEJM), retaining the full breadth of a real diagnostic workup: history, physical findings, laboratory results, diagnostic studies and medical imaging. Each case becomes a simulated encounter. The doctor model receives only the initial presentation of the clinical case and must elicit history, examination findings,
and investigations before committing to a diagnosis. Because the underlying case and patient behaviour are fixed, every model faces the same latent clinical evidence. Performance differences can therefore be attributed to how evidence is acquired, not to variation in the case or the simulator.

We pair each encounter with a static counterpart in which the complete case is given upfront. This turns the comparison into a controlled, within-case contrast between diagnosis given the case in advance and diagnosis through the encounter. Evaluating a diverse set of open- and closed-source LLMs across multiple configurations, we find a substantial and consistent loss of diagnostic accuracy in the interactive setting. This loss cannot be attributed to deficits in medical knowledge or clinical reasoning, since the same models significantly diagnose more accurately the cases when all the evidence is provided in advance. It tracks instead the quality of the information state the model builds during the encounter. Models that pose targeted, discriminating questions reconstruct a more complete clinical picture and reach the correct diagnosis more often.
These results expose a construct-validity problem in current LLM evaluation. Expert-level accuracy on fully specified clinical case descriptions measures inference over curated evidence, not the hypothesis-driven acquisition of that evidence on which clinical diagnosis depends. By separating these two components, Hx2Dx ARENA locates the primary bottleneck of current models in information gathering rather than in reasoning.

Our contributions are fourfold:
\textbf{(1)} We introduce \textbf{Hx2Dx ARENA}, a large-scale, interactive and multimodal benchmark of 1'000 NEJM-derived cases, in which LLMs must progressively elicit history, examination findings, 
and diagnostic investigations before reaching a diagnosis.
\textbf{(2)} We construct a matched static counterpart for every case, enabling a within-case comparison of diagnostic accuracy with and without the burden of information acquisition.
\textbf{(3)} We systematically evaluate a diverse set of open- and closed-source LLMs, compact and large-scale, showing that all models lose diagnostic accuracy in the interactive setting, despite strong performance once the complete case is available.
\textbf{(4)} We identify hypothesis-driven information gathering as the main determinant of interactive performance, specifically the ability to ask questions that discriminate between competing diagnoses. We show that when the right questions are asked, lower-performing models close much of the gap with the strongest ones.

\section{Related Work}

Large language models have rapidly moved from general-purpose text generation
to specialised applications across many domains, including
law~\cite{katz2024gpt,guha2023legalbench},
finance~\cite{wu2023bloomberggpt}, software engineering~\cite{chen2021evaluating,kim2026knowledge},
education~\cite{kasneci2023chatgpt} and scientific
discovery~\cite{romera2024mathematical}, and
medicine has become one of the most active of these domains. Their diagnostic capabilities are now documented at several levels of
the clinical reasoning chain. At the level of clinical information
processing, LLMs can extract symptoms, findings and medications from
unstructured clinical notes~\cite{agrawal2022large}, and summarise clinical documentation with quality comparable to or exceeding that of medical experts~\cite{van2024adapted}. At the level of medical knowledge, they achieve expert-level performance on licensing-style examinations such as MedQA and USMLE~\cite{singhal2023large,singhal2025toward}.
At the level of diagnostic reasoning, they generate differential diagnoses for complex clinicopathological conferences that rival or surpass those of physicians~\cite{mcduff2025towards}, and in some studies they
outperform clinicians on diagnostic reasoning tasks while not significantly improving physicians' performance when used as an aid~\cite{goh2024large,brodeur2024superhuman}. These capabilities span internal medicine \cite{vrdoljak2025evaluating},
radiology \cite{bhayana2024chatbots}, dermatology \cite{goktas2024assessing}, psychiatry \cite{raballo2025diagnosing} and other specialties,
and have motivated a growing number of proposals for LLM-assisted triage, decision support, and second-opinion systems. Across this literature, however, the model typically reasons over a pre-assembled case. The patient's history has already been taken, the examination performed, and the relevant investigations selected by clinicians, so that what is measured is interpretation of available evidence rather than its acquisition.

In fact, clinical LLM evaluation has largely followed two complementary paradigms. Static benchmarks, the vast majority, such as MedHELM~\citep{bedi2026holistic} and HealthBench~\citep{arora2025healthbench} evaluate responses to curated clinical questions or conversations using clinician-authored criteria, while clinician-preference arenas such as MedArena~\citep{wu2026medarena} and MOOVE~\citep{elhassan2026preferred} compare blinded model responses to clinician-authored queries. These approaches provide broad coverage of clinical tasks and, in the case of preference arenas, ecological validity. However, they evaluate the model after the relevant clinical information has already been specified in the prompt. Moreover, recent evidence suggests that human preference may reward surface properties such as verbosity and presentation rather than clinical safety or accuracy~\citep{elhassan2026preferred}. This motivates evaluation settings in which diagnostic performance can be assessed jointly with the process by which information is acquired.

However, diagnosis is inherently sequential: clinicians elicit the history, perform examinations, select investigations, and iteratively update a differential diagnosis under partial observability. Some recent benchmarks and systems therefore evaluate LLMs in interactive settings. AMIE~\citep{tu2025towards} studies diagnostic dialogue with simulated patients, AgentClinic~\citep{schmidgall2026agentclinic} provides multi-agent clinical environments, CRAFT-MD~\citep{johri2025evaluation} evaluates diagnostic
accuracy in conversations with simulated patients compared with the
corresponding clinical vignettes, and MediQ~\citep{li2024mediq} explicitly evaluates the ability to identify missing information and ask targeted questions. Other work has examined end-to-end simulated consultations grounded in clinical records~\citep{niu2026aipatient}. These studies highlight the importance of interactive evaluation and demonstrate that clinical performance can differ substantially once information must be elicited through dialogue. 

We build on this paradigm Hx2Dx ARENA, constructed from 1'000 clinical cases published in the NEJM. Unlike synthetic vignettes or simplified scenarios, every encounter is grounded in the documented course of a real patient. The history reflects the patient's actual narrative, and the final diagnosis was established by the treating team, frequently confirmed by histopathology or other definitive testing. The investigations available to the doctor model are those actually performed in the original workup and are returned exactly as reported. These include physical examination findings, complete laboratory panels with values, units and reference ranges, electrocardiograms, radiographs, cross-sectional imaging, and histopathology, with the images provided as published in the case. No crucial finding is generated by the simulator, so the evidence the model can obtain is the same evidence  available to the clinicians who managed the case. The cases span a broad range of specialties and include rare conditions and atypical presentations, where the discriminating findings are rarely apparent from the presenting complaint and must be actively sought. 

The scaffolding of the arena is designed to emulate a realistic clinical encounter while minimising simulator hallucination. Patient knowledge, investigation results, and the reference diagnosis are separated into distinct artefacts held by distinct agents, so that no single source can disclose the answer. The patient follows an OSCE-style script and answers only what it is asked. The investigation agent retrieves documented results and never generates them. No crucial finding is therefore produced by the simulator, and the evidence the model can obtain is the same evidence that was available to the clinicians who managed the case. Within this setting, we evaluate different classes of LLMs, including proprietary, large open-weight, and smaller open-weight models. Each model is tested on the same cases in both a static setting, where the complete case is provided, and a dynamic setting, where it must be reconstructed through the encounter. To our knowledge, Hx2Dx ARENA is the first benchmark to combine real, multimodal case records at this scale with grounded investigations and a matched static-dynamic evaluation of the same cases. This design lets us systematically characterise  how LLMs perform when the diagnosis depends on progressively eliciting and integrating the history, the examinations, and the investigations of a real clinical encounter. 
It also lets us investigate whether diagnostic failures arise from gathering clinical evidence or from synthesizing it, and thus identify where current models fall short in the diagnostic process.

\begin{figure*}
    \centering
    \includegraphics[width=1\linewidth]{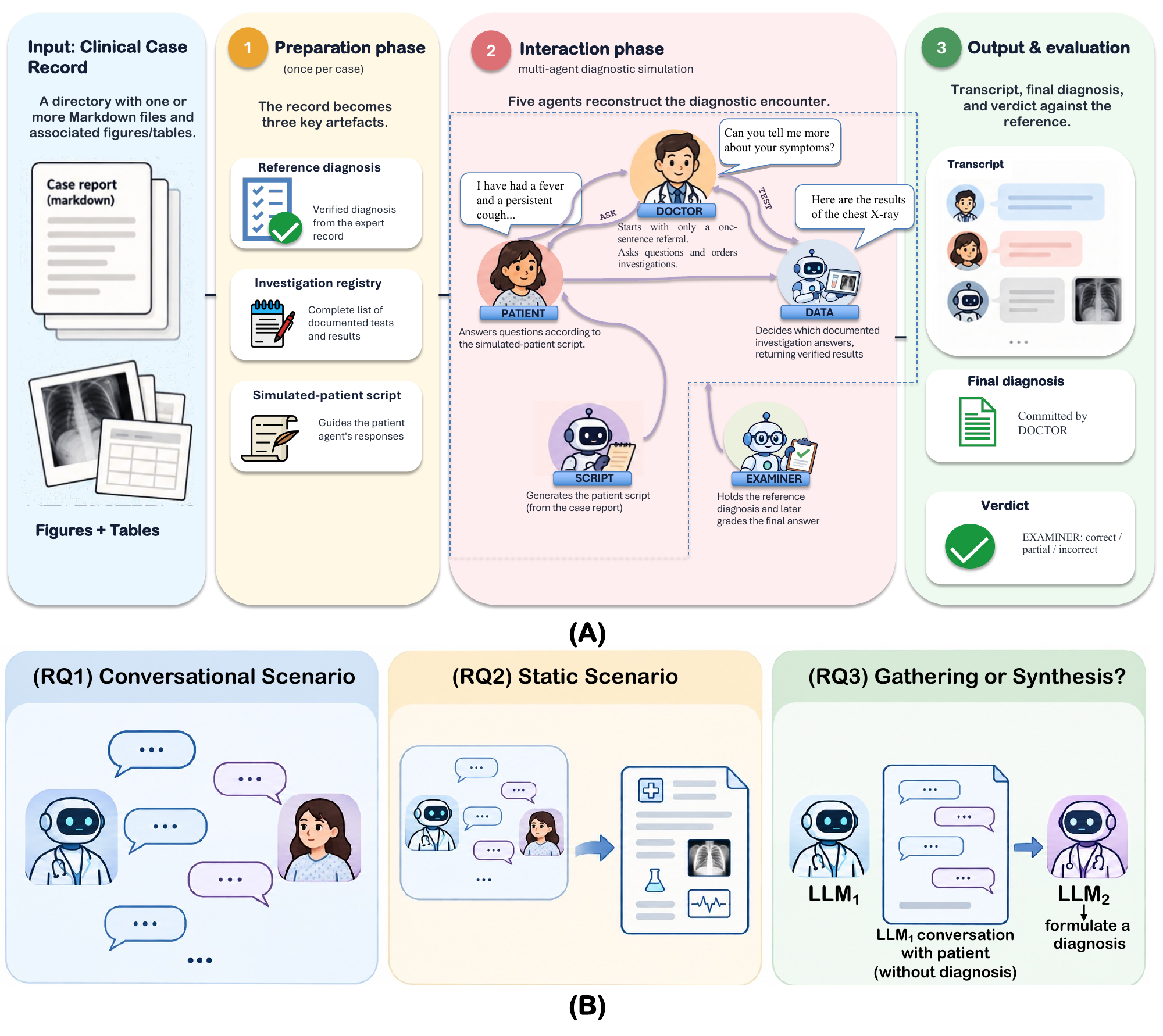}
    \caption{\textbf{(A)} Overview of Hx2Dx ARENA. Each NEJM case is decomposed into a reference diagnosis, an investigation registry, and a simulated-patient script. Starting from a one-sentence referral, the \textsc{Doctor} questions the \textsc{Patient} and requests investigations from \textsc{Data}, which returns only documented results. The \textsc{Examiner} grades the final diagnosis as correct, partial or, incorrect. \textbf{(B)} Visual summary of the three main Research Questions.}
    \label{overview}
\end{figure*}

\section{Research Questions}

The following Research Questions (visually summarised in Figure \ref{overview}\textbf{(B)}) shaped our work:

\begin{itemize}
    \item\textbf{(RQ1)} \textit{(a)} How accurately do LLMs reach the correct diagnosis when they must conduct the encounter themselves, eliciting history, examination findings, and investigations under partial observability? \textit{(b)} How much does the diagnostic performance vary across models?
    \item\textbf{(RQ2)} How much diagnostic accuracy is lost when the same case is reconstructed through the encounter rather than presented as a fully specified record, and is this loss consistent across models and cases?
    \item\textbf{(RQ3)} \textit{(a)} What drives the underperformance of weaker models in the interactive setting? Is it a failure of evidence acquisition, where the discriminating findings are never elicited, or a failure of diagnostic synthesis, where the findings are obtained but misinterpreted? \textit{(b)} Which interventions on the acquisition process improve diagnostic performance?
    \item\textbf{(RQ4)} How does diagnostic performance vary across medical specialties? In which specialties do models struggle most in the interactive setting?
\end{itemize}

\section{Data}
\label{sec:data}

Hx2Dx ARENA is built from clinical case reports published in the \textit{New England Journal of Medicine} between September 1992 and June 2026. These reports are well suited  for an interactive diagnostic benchmark: each documents a real patient encounter in temporal order, from presenting complaint through history, physical examination and diagnostic workup, and closes with a final diagnosis established by the treating team, frequently confirmed by histopathology or other definitive testing. The cases cover a wide range of specialties and include common and rare conditions, typical and atypical presentations, and frequent diagnostic uncertainty, making them a demanding test of clinical reasoning.

We manually excluded all cases in which the patient could not respond verbally to the physician at presentation, for example patients who were intubated, comatose, or otherwise unable to communicate. In such encounters the history is necessarily obtained from relatives, caregivers, or prior medical records. Including them would require additional informants to answer the doctor's questions and would alter the conversational setting, in which the history is elicited directly from the patient. This constraint ensures that every encounter in the arena can be conducted as a dialogue between the
\textsc{Doctor} and a single \textsc{Patient} agent. After filtering, the dataset comprises 1'000 cases.

Each case is stored as a self-contained record. The text of the report is converted to Markdown, preserving its section structure. Tables reporting laboratory and other diagnostic results are converted to Markdown tables, retaining values, units, and reference ranges. All figures embedded in the report, such as radiographs, cross-sectional imaging, electrocardiograms and, histopathology, are saved as PNG files and linked to the corresponding location in the text. This format preserves the multimodal structure of the original publication and serves as input to the preparation phase described in Section~\ref{sec:arena}, where each record is decomposed into the reference diagnosis, the investigation registry, and the simulated-patient script.

\section{Hx2Dx ARENA}
\label{sec:arena}

Hx2Dx ARENA reconstructs the diagnostic encounter underlying each NEJM case as a controlled multi-agent simulation (see Figure~\ref{overview}\textbf{(A)} for a pipeline overview). Its design rests on three principles. \textbf{(1)} the doctor model starts with almost no information (except for a one-sentence referral that includes the patient’s age, sex, and reason for presentation) and must acquire every finding through its own actions (\textit{partial observability}); \textbf{(2)} everything returned to the doctor can be traced to the original case record, so the simulation never introduces findings that the treating physicians did not document (\textit{grounding)}; \textbf{(3)} each case is administered identically to every model, so that differences in outcome can be attributed to the diagnostic strategy of the model under evaluation (\textit{standardisation}).

Each case is provided as a clinical record consisting of one full NEJM report and its associated figures and tables. In a preparation phase, run once per case, the record is decomposed into three artefacts. The \emph{reference diagnosis} is the final diagnosis established in the expert record and serves as ground truth. The \emph{investigation registry} is a structured inventory of every examination and diagnostic study documented in the case together with its result: physical examination findings, laboratory panels, electrocardiography, imaging, histopathology and any other reported test. Each entry keeps its original modality, so a chest radiograph remains the figure published in the case and a laboratory panel remains a table data. The \emph{simulated-patient script} is generated from the case report by a dedicated \textsc{Script} agent. It encodes only what a patient could plausibly report (presenting symptoms and their chronology, past medical history, medications, allergies, family and social history, review of systems) and excludes investigation results and the diagnosis. This decomposition enforces the information asymmetry of clinical practice: history comes from the patient, objective findings from examination and investigations, and neither source discloses the answer.

The encounter itself follows the format of an Objective Structured Clinical Examination (OSCE), in which a candidate assesses a standardised patient whose responses are mostly fixed by a script. We formalise each case as a tuple \begin{equation}   c = \left(x_0,\; \mathcal{S},\; \mathcal{R},\; d^\star\right),   \label{eq:case} \end{equation} where $x_0$ is the one-sentence referral stating age, sex and presenting complaint, $\mathcal{S}$ is the simulated-patient script, $\mathcal{R}$ is the investigation registry, and $d^\star$ is the reference diagnosis. The \textsc{Doctor}, the model under evaluation, observes only $x_0$, comparable to the reason for consultation a physician reads before meeting the patient; $\mathcal{S}$, $\mathcal{R}$ and $d^\star$ remain latent. The encounter is a sequential decision process in which, at each turn $t$, the \textsc{Doctor} selects an action $a_t$ conditioned on the interaction history, \begin{equation}   a_t \sim \pi_\theta\!\left(\cdot \mid h_{t-1}\right), \qquad   h_{t} = \left(x_0, a_1, o_1, \dots, a_{t}, o_{t}\right),   \label{eq:policy} \end{equation} where $\pi_\theta$ is the doctor model and $o_t$ the observation returned by the environment. Each action belongs to the space \begin{equation}   \mathcal{A} = \left\{\texttt{ASK}(q),\; \texttt{TEST}(r),\;   \texttt{FINAL\_DIAGNOSIS}(\hat{d})\right\}.   \label{eq:actions} \end{equation} With $\texttt{ASK}(q)$ the \textsc{Doctor} takes the history, addressing a free-text question $q$ to the \textsc{Patient} agent about, for example, the onset and character of a symptom, exposures, medications or travel. With $\texttt{TEST}(r)$ it asks the \textsc{Data} agent for a specific physical examination component, laboratory test, imaging study or other investigation $r$. With $\texttt{FINAL\_DIAGNOSIS}(\hat{d})$ it commits to a diagnosis $\hat{d}$ and ends the encounter. The environment responds according to \begin{equation}   o_t =   \begin{cases}     P\!\left(q \mid \mathcal{S}\right) & \text{if } a_t = \texttt{ASK}(q),\\[2pt]     \mathcal{R}[r] & \text{if } a_t = \texttt{TEST}(r) \text{ and } r \in \mathcal{R},\\[2pt]     \varnothing & \text{if } a_t = \texttt{TEST}(r) \text{ and } r \notin \mathcal{R},   \end{cases}   \label{eq:observation} \end{equation} where $P(q \mid \mathcal{S})$ is the \textsc{Patient}'s answer constrained by the script, $\mathcal{R}[r]$ is the documented result in its original modality (text or image), and $\varnothing$ denotes an investigation reported as unavailable. The encounter terminates at \begin{equation}   T = \min\left\{\, t : a_t = \texttt{FINAL\_DIAGNOSIS}(\hat{d}) \,\right\},   \qquad T \le T_{\max} = 50,   \label{eq:termination} \end{equation} and if the budget $T_{\max}$ is reached the \textsc{Doctor} is forced to issue $\hat{d}$ conditioned on $h_{T_{\max}}$. The final diagnosis is then graded by the \textsc{Examiner} as $E(\hat{d}, d^\star) \in \{\text{correct}, \text{partial}, \text{incorrect}\}$.  The action space thus reproduces the three components of the diagnostic workup while leaving entirely to the model the order in which they are pursued and how specific each question $q$ or request $r$ is. These choices, encoded in the trajectory $h_T$, constitute the model's information-gathering strategy and are the central object of evaluation.

The behaviour of the environment agents is constrained to preserve this
construct. Following OSCE conventions, the \textsc{Patient} answers only what
it is asked, concisely and in lay language, without volunteering information
beyond the scope of the question. A verbose patient who disclosed large parts
of the script in response to an open question would collapse the interactive
setting into the static one, rewarding models that ask little rather than
models that ask well. Concise answers ensure that discriminating findings
emerge only in response to targeted questions, as they do at the bedside. 

The \textsc{Data} agent matches each request against the investigation registry and returns the documented result in its original modality. If the requested investigation was not performed in the original case, the \textsc{Doctor} is told it is unavailable. Because \textsc{Data} retrieves results and never generates them, every finding the doctor sees was actually obtained by the treating team, which prevents simulator hallucination from contaminating the evaluation. Both \textsc{Patient}, \textsc{Examiner}, and \textsc{Data} are instantiated with \texttt{Gemini~3.5~Flash~Lite} with temperature 0 to force deterministic outputs. The encounter is capped at \textbf{50} turns of questions and investigation requests. Once the budget is exhausted the \textsc{Doctor} is forced to issue a final diagnosis on the evidence gathered so far. The cap reflects the bounded nature of a real consultation and prevents exhaustive enumeration of tests from substituting for hypothesis-driven reasoning.

Each encounter yields a complete transcript, the doctor's final diagnosis and a verdict. As already stated, the \textsc{Examiner} agent, which holds the reference diagnosis, grades the final answer as correct, partially correct, or incorrect: 
\begin{enumerate}
    \item Correct for the reference diagnosis or a clinically equivalent formulation;
    \item Partial Correct for a diagnosis that captures the right disease category or mechanism without the specificity of the reference;
    \item Incorrect otherwise.
\end{enumerate}
In our primary analyses we count partially correct diagnoses as successes, for two reasons. First, NEJM reference diagnoses are often highly specific, down to a histological subtype, a causative organism, or a genetic variant, and this level of specificity is typically established only by definitive tests performed late in the workup, such as biopsy, culture, or molecular testing. Second, a diagnosis that identifies the correct disease category or pathophysiological mechanism is clinically actionable: it directs the patient to the appropriate specialist, the appropriate confirmatory investigations and, often, the appropriate initial management. In clinical practice this corresponds to a correct working diagnosis, from which the definitive diagnosis follows. Strictly correct diagnoses are numerically reported separately (see Table \ref{simulation_verdicts} in Appendix). The transcript also allows the encounter itself to be analysed: which questions were asked, which investigations were requested and in what order, and how much of the documented evidence was recovered.

\textbf{Static counterpart.}
For each case we construct a static counterpart that removes the burden of information acquisition while keeping the available evidence identical. The \textsc{Doctor} receives, in a single prompt, the referral $x_0$ together with the full content of the simulated-patient script $\mathcal{S}$ and of the investigation registry $\mathcal{R}$, including all images in their original modality, and is asked to provide a final diagnosis: \begin{equation}   \hat{d}_{\mathrm{static}} = \pi_\theta\!\left(x_0,\, \mathcal{S},\,   \mathcal{R}\right).   \label{eq:static} \end{equation} The reference diagnosis $d^\star$ and the diagnostic discussion of the NEJM report are excluded, and $\hat{d}_{\mathrm{static}}$ is graded by the same \textsc{Examiner}. Since $(\mathcal{S}, \mathcal{R})$ is exactly the evidence the \textsc{Doctor} could in principle obtain through the encounter, the static setting corresponds to an interaction in which all latent information has been elicited. Any difference between $E(\hat{d}_{\mathrm{static}}, d^\star)$ and $E(\hat{d}, d^\star)$ can therefore be attributed to evidence acquisition rather than to differences in the underlying case. The static setting therefore provides a natural upper bound on each model's performance in the arena, determined solely by its capacity for diagnostic synthesis.

\subsection{Models Used}

We evaluated seven large language models as the \textsc{DOCTOR} agent. We restricted the selection to natively multimodal models, so that every agent could inspect the clinical images attached to the case records, and we chose them to span a range of sizes and of access models, including both proprietary models served through  APIs provider and open-weight models. Table~\ref{models} summarises their characteristics. Four models have built-in reasoning (``thinking'') capabilities, which we left enabled at their default settings in all experiments, whereas the three smaller open-weight models do not. Following common practice\cite{guha2023legalbench,chen2023put}, all models were run with temperature set to 0.

\begin{table}[h]
\centering
\rowcolors{2}{gray!18}{white}
\begin{tabular}{lcccc}
\toprule
\textbf{Model} & \textbf{Access} & \textbf{Parameters} & \textbf{Multimodal} & \textbf{Built-in thinking} \\
\midrule
GPT-6 Luna            & Proprietary & --    & \checkmark & Yes \\
Gemini 3.5 Flash Lite & Proprietary & --    & \checkmark & Yes \\
DeepSeek V4.1 Flash   & Open-weight & 763B  & \checkmark & Yes \\
GLM 5.3 Flash         & Open-weight & 321B  & \checkmark & Yes \\
Llama 4 Scout         & Open-weight & 109B  & \checkmark & No  \\
Gemma 3 27B IT        & Open-weight & 27B   & \checkmark & No  \\
Ministral 8B          & Open-weight & 8B    & \checkmark & No  \\
\bottomrule
\end{tabular}
\caption{Models evaluated as the physician agent. Parameter counts are reported as total  parameters; ``-'' denotes undisclosed
values.}
\label{models}
\end{table}

\section{Results} \label{sec:results}

We evaluated seven LLMs on the 1'000 cases of Hx2Dx ARENA in both the conversational and the static setting.

Diagnostic accuracy is the proportion of cases graded correct or partially correct by the \textsc{Examiner}. Uncertainty is reported as 95\% bootstrap confidence intervals (10'000 resamples). We report results for each research question in turn.

\begin{figure*}
    \centering
    \includegraphics[width=1\linewidth]{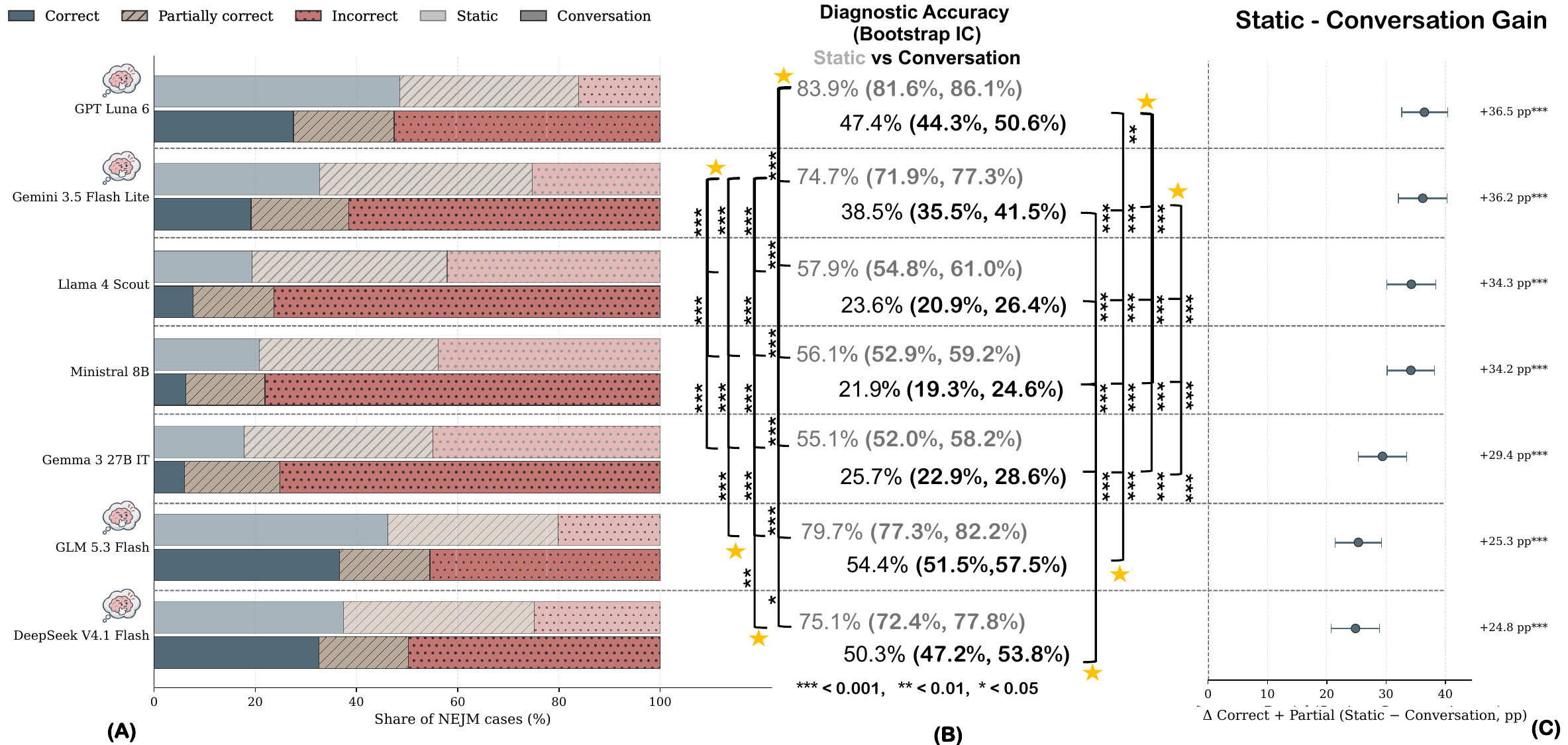}
    \caption{Diagnostic performance of seven LLMs on 1,000 NEJM clinical cases in the static and conversational settings. \textbf{(A)} Distribution of judge verdicts (Correct, Partially correct, Incorrect) as a percentage of cases. For each model, the upper (light) bar shows the static setting, where a description of case report is given, and the lower (dark) bar shows the conversational setting, where the model has to gather information by questioning a simulated patient. \textbf{(B)} Diagnostic accuracy, defined as Correct + Partially correct, with 95\% bootstrap confidence intervals (10'000 resamples), for the static (\textcolor{gray}{gray}) and conversational (black) settings. Brackets mark pairwise comparisons between models within the static setting (left) and within the conversational setting (right). Each bracket group compares one reference model, marked with a gold star (\goldstar), against each of the other models. We used exact McNemar tests with Benjamini–Hochberg FDR correction ($^{***} p < 0.001, ^{**} p < 0.01, ^{*} p < 0.05$); only significant comparisons are shown. \textbf{(C)} Accuracy loss when moving from the static to the conversational setting (Static - Conversation, in percentage points), with 95\% bootstrap CIs. Stars give the significance of the difference (McNemar tests with Benjamini–Hochberg FDR correction). Models are ordered by accuracy loss.}
    \label{res}
\end{figure*}

\subsection{RQ1: Diagnostic performance in the conversational setting}  

When models had to reconstruct the case through the encounter, diagnostic accuracy varied widely across models (Figure~\ref{res}\textbf{(A-B)}, darker
bars in panel A and black values in panel B). GLM~5.3 Flash achieved the highest accuracy, reaching a correct or partially correct diagnosis in 54.4\% of cases. It was followed by DeepSeek~V4.1 Flash (50.3\%), GPT Luna~6 (47.4\%) and Gemini~3.5 Flash Lite (38.5\%). The three remaining models formed a distinctly weaker group: Gemma~3 27B IT (25.7\%), Ministral~8B (21.9\%) and Llama~4 Scout (23.6\%). Proprietary and large open-weight models thus significantly outperformed the smaller open-weight models, with a difference of 32.5 percentage points between the best and the worst model.

\subsection{RQ2: From the static to the conversational setting}  

In the static setting, where the complete case was available upfront, all models performed substantially better (Figure~\ref{res}\textbf{(A-B)}, lighter bars in panel A and grey values in panel B). GPT Luna~6 achieved the highest static accuracy (83.9\%), followed by GLM~5.3 Flash (79.7\%), DeepSeek~V4.1 Flash (75.1\%) and Gemini~3.5 Flash Lite (74.7\%). Llama~4 Scout (57.9\%), Ministral~8B (56.1\%) and Gemma~3 27B IT (55.1\%) again formed the lower tier.  Every model lost accuracy when moving from the static to the conversational setting, and every loss was statistically significant (Figure~\ref{res}\textbf{(C)}). 
Since the static setting provides an upper bound determined solely by diagnostic synthesis, we quantify this loss for each model $m$ as \begin{equation}   \Delta_m = \mathrm{Acc}^{\mathrm{static}}_m - \mathrm{Acc}^{\mathrm{conv}}_m,   
\end{equation} which we refer to as the \textit{gathering loss}: the share of accuracy lost because the model must elicit the clinical evidence through the encounter rather than receive it complete.
The magnitude of this loss, however, varied considerably. GPT Luna~6 showed the largest drop in percentage point (pp) (36.5~pp), followed by Gemini~3.5 Flash Lite (36.2~pp), Llama~4 Scout (34.3~pp), Ministral~8B (34.2~pp) and Gemma~3 27B IT (29.4~pp). DeepSeek~V4.1 Flash (24.8~pp) and GLM~5.3 Flash (25.3~pp) were the most robust to the interactive setting. Because losses differed across models, the ranking was not preserved across settings. Most notably, GPT Luna~6, the strongest model when given the full case, fell to third place in the conversational setting, behind GLM~5.3 Flash and DeepSeek~V4.1 Flash. 
Note that in the static setting, the model is presented with the entire content of the case and therefore with a complete clinical picture, including investigations that may in themselves point to the diagnosis. 

\begin{figure*}
    \centering
    \includegraphics[width=1\linewidth]{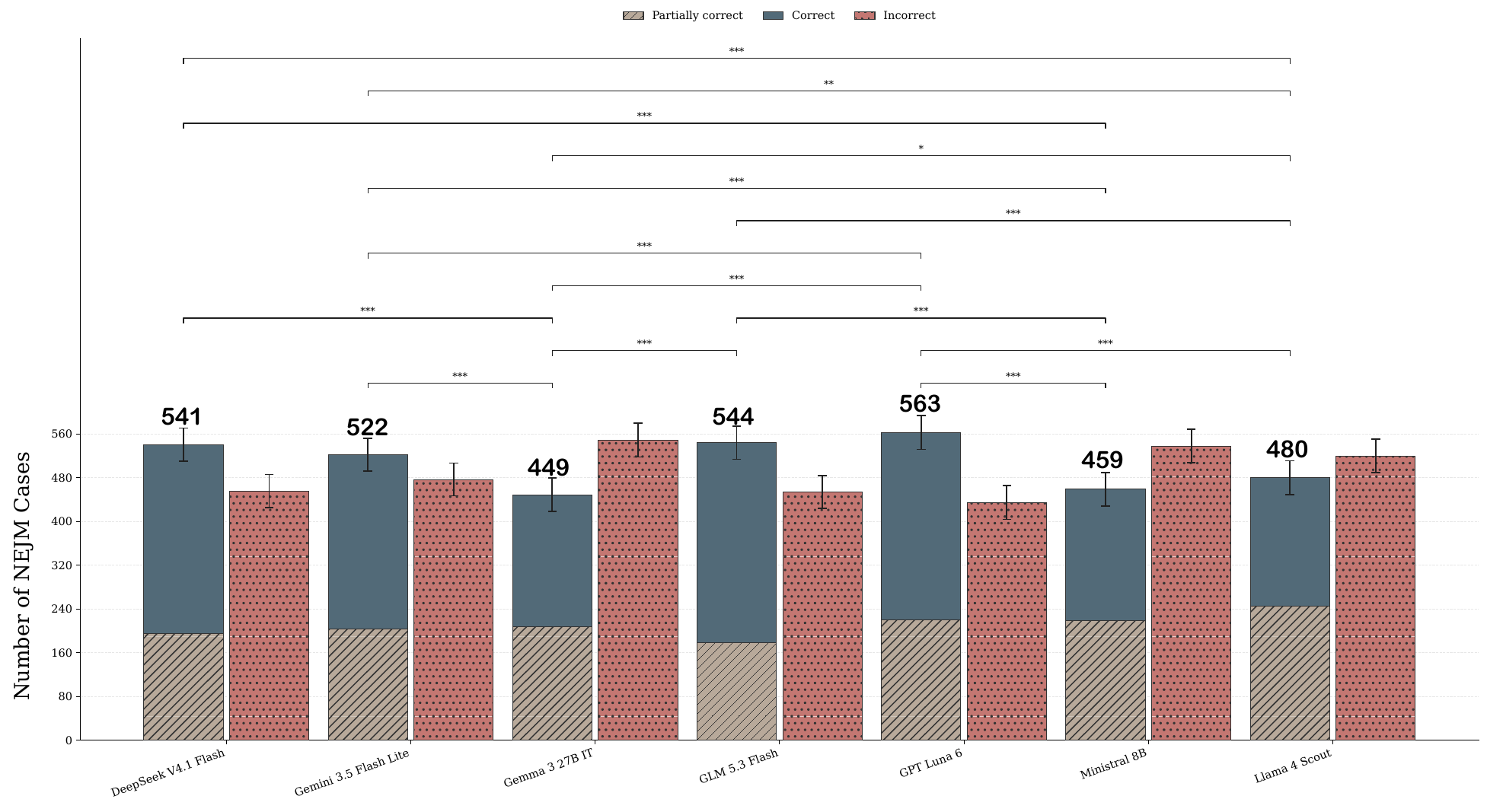}
    \caption{Diagnostic accuracy when all models receive the same conversation transcripts from GLM 5.3 Flash, the best performing model. For each model, the left stacked bar shows the number of cases judged Correct or Partially correct, with the total above the bar. The right bar shows Incorrect cases. Error bars are 95\% bootstrap confidence intervals (10'000 resamples). Brackets mark significant pairwise differences in accuracy (Correct + Partially correct), from exact McNemar tests with Benjamini–Hochberg FDR correction ($^{***} p < 0.001, ^{**} p < 0.01, ^{*} p < 0.05$). }
    \label{replay}
\end{figure*}

\subsection{RQ3: Evidence gathering or diagnostic synthesis?}  

To separate the two sources of error, we decoupled information gathering from diagnosis. We took the conversation transcripts produced by GLM~5.3 Flash, the best-performing model in the conversational setting, removed its final diagnosis, and asked every model to formulate a diagnosis from these transcripts (Figure~\ref{replay}). 

For each case, the transcript was converted back into the exact message sequence the original physician agent had received: the same system prompt and opening referral, followed by its questions and test orders, the simulated patient’s answers, the test results, and any study images together with GLM’s written readings of them. The sequence was injected into the target model’s context as a prefilled multi-turn history, with GLM’s actions placed in the assistant role, so that each model continued the encounter as if it had conducted it. Only the visible conversation was transferred; GLM’s internal reasoning traces were not. The target model was then prompted to end the encounter and commit to a final answer in the same \texttt{DIAGNOSIS/DIFFERENTIAL/REASONING} format. Decoding was greedy (temperature 0), and answers were graded by the same LLM judge used in the conversational setting. 

Because the information available is held constant across models, differences in accuracy in this setting reflect diagnostic reasoning alone, independently of each model’s information-gathering strategy.

Given the GLM~5.3 Flash transcripts, accuracy rose for every other model. The largest gains were observed for the weakest interviewers: Llama~4 Scout improved from 23.6\% to 48.0\% (+24.4~pp), Ministral~8B from 21.9\% to 45.9\% (+24.0~pp) and Gemma~3 27B IT from 25.7\% to 44.9\% (+19.2~pp). Gemini~3.5 Flash Lite improved from 38.5\% to 52.2\% (+13.7~pp), GPT Luna~6 from 47.4\% to 56.3\% (+8.9~pp), and DeepSeek~V4.1 Flash from 50.3\% to 54.1\% (+3.8~pp). GPT Luna~6 diagnosed more accurately from the GLM~5.3 Flash transcripts than GLM~5.3 Flash itself (56.3\% vs 54.4\%, with no statistical difference), indicating that its conversational deficit lies, once again, in how it gathers evidence rather than in how it interprets it.  These results answer \textbf{(RQ3)a.}: the underperformance of models in the interactive setting is predominantly a failure of evidence acquisition. When the discriminating findings are elicited for them, models that diagnose in only one case out of five on their own reach accuracies within a few percentage points of the strongest models. They also answer \textbf{(RQ3)b.}: improving the acquisition process, here by substituting the questioning of a stronger interviewer, is sufficient to recover most of the conversational loss for weaker models.  The intervention did not, however, close the gap to the static setting. Even with the transcripts of the best interviewer, accuracy remained 9.9 to 27.6~pp below each model's static accuracy, with the largest residual gap for GPT Luna~6 (56.3\% vs 83.9\%). Part of the evidence available in the full case therefore remained unelicited  by the best-performing model.  

\begin{figure}
    \centering
    \includegraphics[width=1\linewidth]{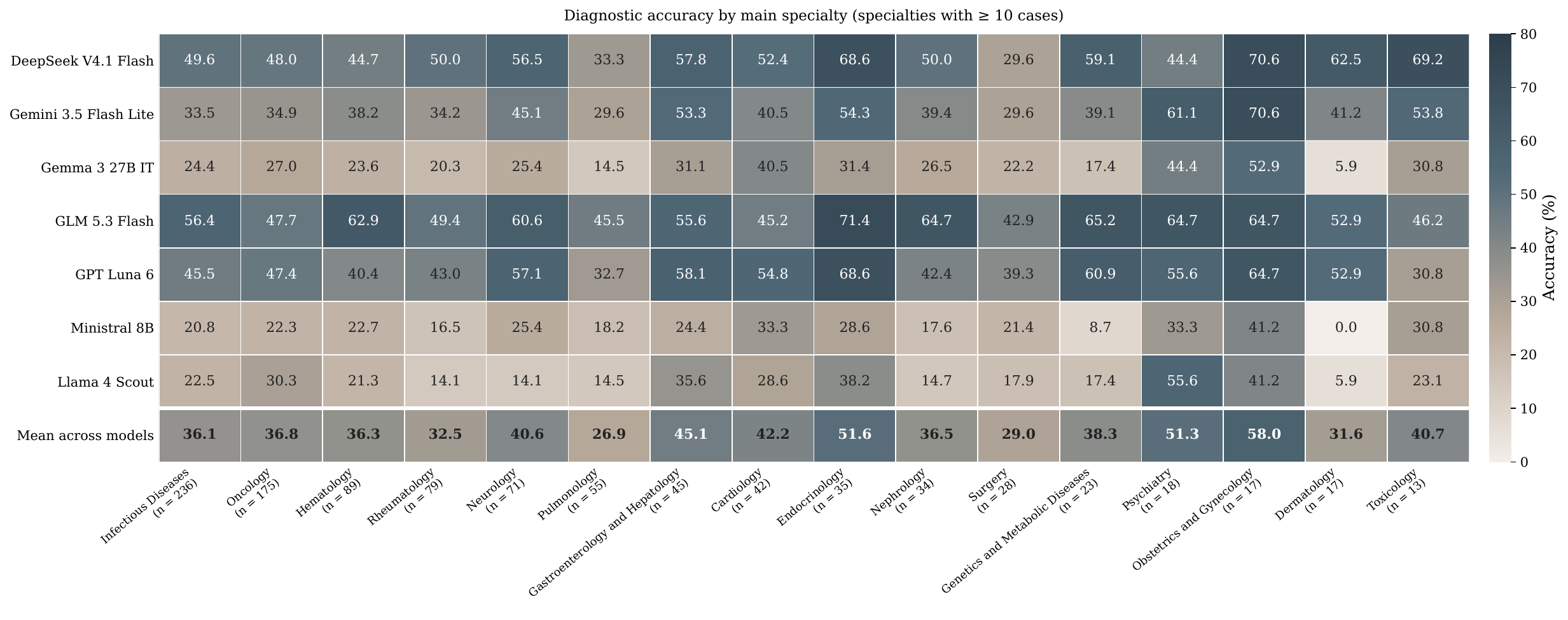}
    \caption{Diagnostic accuracy in the conversational setting, stratified by main clinical specialty. Each cell gives the percentage of cases judged Correct or Partially correct for one model (rows) and one specialty (columns). Only specialties with at least 10 cases are shown; the number of cases (n) is given below each specialty name. Specialties are ordered by number of cases. The bottom row is the unweighted mean across the seven models.}
    \label{spec}
\end{figure}

\subsection{RQ4: Diagnostic performance across specialties}  

Figure~\ref{spec} reports conversational accuracy stratified by the main clinical specialty of each case, for the 16 specialties with at least 10 cases (977 of 1'000 cases); each case was assigned a single primary specialty by \texttt{Gemini~3.5 Flash Lite} 
based on its reference diagnosis and chosen from a predefined list of 22 specialties. The case mix reflects the NEJM corpus and is dominated by infectious diseases ($n=236$), oncology ($n=175$), hematology ($n=89$), rheumatology ($n=79$) and neurology ($n=71$).  Averaged across models, pulmonology was the most difficult specialty, with an accuracy of 26.9\% ($n=55$). It was the only specialty in which six of the seven models remained below 35\%. Surgery (29.0\%, $n=28$), rheumatology (32.5\%, $n=79$) and dermatology (31.6\%, $n=17$) were also below the overall mean. In dermatology, two models, Gemma~3 27B IT and Llama 4 Scout, reached the correct diagnosis in only one of 17 cases (5.9\%), while Ministral 8B none. The three largest specialties, infectious diseases (36.1\%), oncology (36.8\%) and hematology (36.3\%), showed very similar accuracies. The highest accuracies were observed in obstetrics and gynecology (58.0\%, $n=17$), endocrinology (51.6\%, $n=35$), and  psychiatry (51.3\%, $n=18$). Note that these are among the smallest strata, and their estimates are therefore not particularly robust.  The relative difficulty of specialties was broadly consistent across models. The specialty profile, however, differed between the strongest and weakest interviewers. GLM~5.3 Flash maintained accuracy above 42\% in every specialty, whereas the three weakest models fell below 20\% in most specialties.

\section{Discussion}
\label{sec:discussion}

Hx2Dx ARENA separates two components of diagnosis that static evaluation conflates: the synthesis of a complete clinical picture and the gathering of the evidence from which that picture is built. Across seven different LLMs and 1'000 NEJM cases, every model lost diagnostic accuracy when it had to conduct the encounter itself, with losses ranging from 24.8 to 36.5 percentage points. The transcript-transfer experiment \textbf{(RQ3)}, indicates that most of this loss arises in gathering rather than in synthesis. When the evidence elicited by the strongest interviewer was made available to all models, weaker models recovered up to three quarters of their conversational loss, and the spread between models narrowed from 32.5 to 11.4 percentage points. Current models therefore appear considerably more capable of reasoning over a well-assembled case than of assembling it.

From a clinical perspective, this pattern is consistent with what is known about diagnostic expertise. The history alone establishes the diagnosis in the majority of patients~\cite{hampton1975relative}, and the skill of the experienced clinician lies less in recognising a disease from a complete description than in the hypothesis-driven selection of questions, examination manoeuvres and investigations that discriminate between competing diagnoses. Pertinent negatives must be sought as deliberately as pertinent positives, and the decision to stop investigating is itself a clinical judgement. Because the \textsc{Patient} answers only what it is asked, as a standardised patient would in an OSCE exam, failures in this process have direct consequences in the arena. A model that commits to a diagnosis before the discriminating findings are obtained reproduces premature closure, and a model that pursues an early hypothesis without seeking the evidence that would refute it reproduces anchoring. In both cases the relevant information was present in the case and would have been returned on request.  

\begin{figure}
    \centering
    \includegraphics[width=1\linewidth]{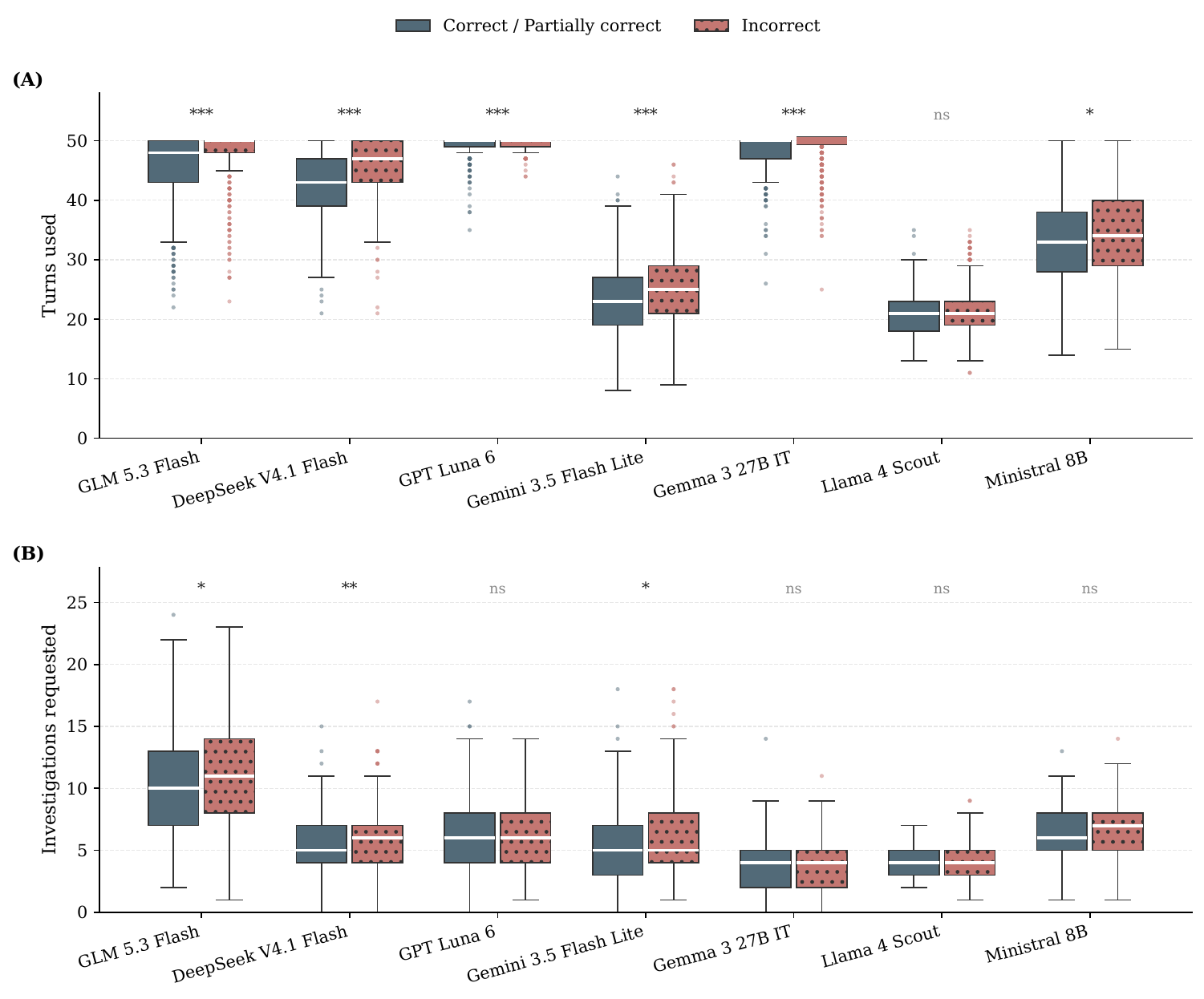}
    \caption{Process metrics and diagnostic accuracy in the conversational setting. \textbf{(A)} Turns used and \textbf{(B)} investigations requested per case, for correctly (Correct or Partially correct) and incorrectly diagnosed cases. Boxes show the median and IQR, and dots are outliers. Thick bars mark distributions collapsed at the 50-turn limit. Within each model, the two groups were compared with Mann–Whitney U tests, with Benjamini–Hochberg FDR correction across all 14 tests ($^{***} p < 0.001, ^{**} p < 0.01, ^* p < 0.05$; ns, not significant). }
    \label{additional}
\end{figure}

From a modelling perspective, the conversational setting turns diagnosis into sequential decision-making under partial observability, in which each action should be chosen for its expected information gain with respect to the current differential. Most training signals available to current models do not target this capability. Medical question answering, instruction tuning on single-turn exchanges, and outcome-based reinforcement learning on fully specified problems all reward the correctness of a final answer given complete inputs, not the quality of the queries that produce those inputs. Our results suggest that these two capabilities are largely dissociable. GPT Luna~6 achieved the highest static accuracy and, given the GLM~5.3 Flash transcripts, diagnosed more accurately than GLM~5.3 Flash itself (56.3\% vs 54.4\%), yet it ranked third when it had to gather the evidence on its own. Strong synthesis therefore does not imply effective gathering, and improvements on static benchmarks need not transfer to interactive diagnosis. This is in line with work showing that models rarely succeed when the information available to them is insufficient~\cite{li2024mediq}.

The transfer experiment also shows that gathering, and not only synthesis, limits the strongest models. With the transcripts of the best interviewer, every model remained below its static accuracy, and the models with the greatest capacity for synthesis are those most constrained by an incomplete picture. In this sense, the ceiling of conversational diagnosis is currently set by the interviewer.

The results conditioned on the different specialties points in the same direction. Pulmonology, where the discriminating evidence often lies in imaging, pulmonary function testing and occupational or environmental exposures, was the most difficult specialty. Dermatology, where the diagnosis typically rests on inspection of the lesion, also proved difficult for several models. Both require the model to recognise that a specific examination or image is worth obtaining before it can be interpreted. Psychiatry and endocrinology, where history and targeted laboratory testing follow more directly from the presenting complaint, showed higher accuracy, as did obstetrics and gynecology, where sex and pregnancy status in the referral already constrain the differential. These results suggest that difficulty in the interactive setting depends on where the discriminating evidence is located and on how far it lies from the initial presentation. 

Process measures support this interpretation (Figure~\ref{additional}). Within models, encounters ending in an incorrect diagnosis were not shorter than those ending in a correct one. For most models they lasted slightly longer and involved slightly more investigations, although the differences were small, typically one to a few turns and about one investigation. Between models, the length of the encounter did not track accuracy either. GPT Luna~6 and Gemma~3 27B IT both used nearly the entire 50-turn budget, yet differed by almost 20 percentage points in accuracy, while Gemini~3.5 Flash Lite and Llama~4 Scout both concluded after about 20--25 turns with similarly divergent results. GLM~5.3 Flash, the most accurate model, requested markedly more investigations than any other model. Diagnostic failure therefore does not reflect insufficient investigative effort. Longer and more test-intensive encounters more often accompany diagnostic uncertainty or an unfocused, shotgun approach to the workup, and what distinguishes successful encounters is the discriminating value of the questions and investigations rather than their number.

These findings have implications for both evaluation and system design. For evaluation, accuracy on fully specified cases measures synthesis and should not be read as evidence of competence in conducting a clinical encounter. This distinction matters for applications in which an LLM interacts with the patient before a clinician does, such as pre-consultation history taking, symptom assessment or triage, where the model's questions determine what evidence will ever be available. 

For system design, the separability of gathering and synthesis suggests two directions. Gathering can be optimised directly, for instance by rewarding actions according to the diagnostic information they yield rather than only rewarding the final diagnosis. Alternatively, it can be delegated to a dedicated agent whose transcripts are then passed to a separate diagnostic model, an architecture that our transfer experiment approximates.

\section{Conclusion \& Future Works}

Hx2Dx ARENA shows that the distance between current LLMs and the diagnostic encounter lies less in synthesising clinical evidence than in gathering it. Models that perform well when the complete case is available lose a substantial share of their accuracy when they must take the history and choose the workup themselves, and much of this loss is recovered when the right questions are asked for them. Hypothesis-driven information gathering is therefore a distinct capability that static evaluations do not measure and that the safe use of LLMs in clinical encounters will require. 

Our evaluation covers seven models. Extending it to a broader range of architectures, including larger reasoning models and models specialised for medicine, will be necessary to establish how general these findings are. In its current form, Hx2Dx ARENA serves as a benchmark of clinical reasoning that assesses whether a model can conduct the encounter, from history taking through examination to the selection of investigations, and not only whether it can recognise a diagnosis from a completed case. 

Additionally, NEJM cases are selected for their educational value and tend to feature complex or uncommon presentations. This makes them a demanding test of diagnostic reasoning, but absolute accuracies may differ in routine clinical settings. Because the cases span more than three decades, some may have been encountered during model training. The interactive setting mitigates this, since the case is never presented in full, and an analysis stratified by publication date could quantify any residual effect. The current version of the arena focuses on patients able to communicate verbally, on encounters conducted in English and in text, and on a budget of 50 turns; extending it to other languages, to encounters involving relatives or caregivers, and to richer interaction formats is a natural next step.

Because every encounter is grounded in a documented case and yields a full transcript, the arena can also go beyond evaluation and serve as an environment for improving models. Encounters could provide feedback on the discriminating value of each question or investigation, training signals could reward efficient and targeted workups rather than final answers alone, and the transcripts could support the analysis of recurrent reasoning errors such as premature closure or anchoring. In this sense, Hx2Dx ARENA offers both a measure of how far current models are from conducting a diagnostic encounter and a way to narrow that distance.






\begin{appendices}

\section{Distribution of verdicts} \label{secA1}  

Table~\ref{simulation_verdicts} reports the distribution of the \textsc{Examiner}'s verdicts in the conversational setting. Considering only strictly correct diagnoses, GLM~5.3 Flash again achieved the highest accuracy (36.6\%), followed by DeepSeek~V4.1 Flash (32.7\%), GPT Luna~6 (27.6\%) and Gemini~3.5 Flash Lite (19.2\%). The three smaller open-weight models reached the exact diagnosis in fewer than 8\% of encounters: Llama~4 Scout (7.7\%), Ministral~8B (6.3\%) and Gemma~3 27B IT (6.2\%). The ordering of the four strongest models is therefore unchanged under the stricter criterion.  The proportion of partially correct diagnoses was comparatively stable across models, ranging from 15.6\% to 19.8\%, so differences in overall accuracy are driven mainly by strictly correct diagnoses. The composition of successful encounters, however, differed markedly between groups of models. For GLM~5.3 Flash and DeepSeek~V4.1 Flash, about two thirds of correct or partially correct diagnoses were strictly correct. For the smaller open-weight models, by contrast, between 67\% and 76\% of successful encounters were only partially correct. These models more often identified the right disease category or mechanism without reaching the specific diagnosis, which typically depends on the discriminating findings that a more complete workup would have elicited.

\begin{table}[t]
\centering
\small
\renewcommand{\arraystretch}{1.4}
\setlength{\tabcolsep}{10pt}
\begin{tabular}{l r r r}
\toprule
\textbf{Model}
  & \textcolor{cCorrect}{\textbf{Correct}}
  & \textcolor{cPartial!75!black}{\textbf{Partially correct}}
  & \textcolor{cIncorrect}{\textbf{Incorrect}} \\
\midrule
GLM 5.3 Flash         & \cellcolor{cCorrect!35}\textbf{366}  & \cellcolor{cPartial!33}178  & \cellcolor{cIncorrect!22}456  \\
DeepSeek V4.1 Flash   & \cellcolor{cCorrect!28}327 & \cellcolor{cPartial!34}176 & \cellcolor{cIncorrect!25}497  \\
GPT Luna 6            & \cellcolor{cCorrect!24}276  & \cellcolor{cPartial!33}\textbf{198}  & \cellcolor{cIncorrect!27}526 \\
Gemini 3.5 Flash Lite & \cellcolor{cCorrect!21}192  & \cellcolor{cPartial!35}193  & \cellcolor{cIncorrect!28}615  \\
Gemma 3 27B IT        & \cellcolor{cCorrect!10}62  & \cellcolor{cPartial!31}195 & \cellcolor{cIncorrect!34}743  \\
Ministral 8B          & \cellcolor{cCorrect!11}63 & \cellcolor{cPartial!26}156  & \cellcolor{cIncorrect!35}\textbf{781} \\
Llama 4 Scout         & \cellcolor{cCorrect!10}77  & \cellcolor{cPartial!27}159  & \cellcolor{cIncorrect!35}764  \\
\bottomrule
\end{tabular}
\caption{Judge verdicts in the conversational setting (correct, partially correct, and incorrect). 
Cell shading is proportional to the percentage within each column; the highest value per
column is in bold. }
\label{simulation_verdicts}
\end{table}




\end{appendices}


\bibliography{sn-bibliography}


\begin{thebibliography}{29}
\ifx \bisbn   \undefined \def \bisbn  #1{ISBN #1}\fi
\ifx \binits  \undefined \def \binits#1{#1}\fi
\ifx \bauthor  \undefined \def \bauthor#1{#1}\fi
\ifx \batitle  \undefined \def \batitle#1{#1}\fi
\ifx \bjtitle  \undefined \def \bjtitle#1{#1}\fi
\ifx \bvolume  \undefined \def \bvolume#1{\textbf{#1}}\fi
\ifx \byear  \undefined \def \byear#1{#1}\fi
\ifx \bissue  \undefined \def \bissue#1{#1}\fi
\ifx \bfpage  \undefined \def \bfpage#1{#1}\fi
\ifx \blpage  \undefined \def \blpage #1{#1}\fi
\ifx \burl  \undefined \def \burl#1{\textsf{#1}}\fi
\ifx \doiurl  \undefined \def \doiurl#1{\url{https://doi.org/#1}}\fi
\ifx \betal  \undefined \def \betal{\textit{et al.}}\fi
\ifx \binstitute  \undefined \def \binstitute#1{#1}\fi
\ifx \binstitutionaled  \undefined \def \binstitutionaled#1{#1}\fi
\ifx \bctitle  \undefined \def \bctitle#1{#1}\fi
\ifx \beditor  \undefined \def \beditor#1{#1}\fi
\ifx \bpublisher  \undefined \def \bpublisher#1{#1}\fi
\ifx \bbtitle  \undefined \def \bbtitle#1{#1}\fi
\ifx \bedition  \undefined \def \bedition#1{#1}\fi
\ifx \bseriesno  \undefined \def \bseriesno#1{#1}\fi
\ifx \blocation  \undefined \def \blocation#1{#1}\fi
\ifx \bsertitle  \undefined \def \bsertitle#1{#1}\fi
\ifx \bsnm \undefined \def \bsnm#1{#1}\fi
\ifx \bsuffix \undefined \def \bsuffix#1{#1}\fi
\ifx \bparticle \undefined \def \bparticle#1{#1}\fi
\ifx \barticle \undefined \def \barticle#1{#1}\fi
\bibcommenthead
\ifx \bconfdate \undefined \def \bconfdate #1{#1}\fi
\ifx \botherref \undefined \def \botherref #1{#1}\fi
\ifx \url \undefined \def \url#1{\textsf{#1}}\fi
\ifx \bchapter \undefined \def \bchapter#1{#1}\fi
\ifx \bbook \undefined \def \bbook#1{#1}\fi
\ifx \bcomment \undefined \def \bcomment#1{#1}\fi
\ifx \oauthor \undefined \def \oauthor#1{#1}\fi
\ifx \citeauthoryear \undefined \def \citeauthoryear#1{#1}\fi
\ifx \endbibitem  \undefined \def \endbibitem {}\fi
\ifx \bconflocation  \undefined \def \bconflocation#1{#1}\fi
\ifx \arxivurl  \undefined \def \arxivurl#1{\textsf{#1}}\fi
\csname PreBibitemsHook\endcsname

\bibitem[\protect\citeauthoryear{Singhal et~al.}{2023}]{singhal2023large}
\begin{barticle}
\bauthor{\bsnm{Singhal}, \binits{K.}},
\bauthor{\bsnm{Azizi}, \binits{S.}},
\bauthor{\bsnm{Tu}, \binits{T.}},
\bauthor{\bsnm{Mahdavi}, \binits{S.S.}},
\bauthor{\bsnm{Wei}, \binits{J.}},
\bauthor{\bsnm{Chung}, \binits{H.W.}},
\bauthor{\bsnm{Scales}, \binits{N.}},
\bauthor{\bsnm{Tanwani}, \binits{A.}},
\bauthor{\bsnm{Cole-Lewis}, \binits{H.}},
\bauthor{\bsnm{Pfohl}, \binits{S.}}, \betal:
\batitle{Large language models encode clinical knowledge}.
\bjtitle{Nature}
\bvolume{620}(\bissue{7972}),
\bfpage{172}--\blpage{180}
(\byear{2023})
\end{barticle}
\endbibitem

\bibitem[\protect\citeauthoryear{Singhal et~al.}{2025}]{singhal2025toward}
\begin{barticle}
\bauthor{\bsnm{Singhal}, \binits{K.}},
\bauthor{\bsnm{Tu}, \binits{T.}},
\bauthor{\bsnm{Gottweis}, \binits{J.}},
\bauthor{\bsnm{Sayres}, \binits{R.}},
\bauthor{\bsnm{Wulczyn}, \binits{E.}},
\bauthor{\bsnm{Amin}, \binits{M.}},
\bauthor{\bsnm{Hou}, \binits{L.}},
\bauthor{\bsnm{Clark}, \binits{K.}},
\bauthor{\bsnm{Pfohl}, \binits{S.R.}},
\bauthor{\bsnm{Cole-Lewis}, \binits{H.}}, \betal:
\batitle{Toward expert-level medical question answering with large language models}.
\bjtitle{Nature medicine}
\bvolume{31}(\bissue{3}),
\bfpage{943}--\blpage{950}
(\byear{2025})
\end{barticle}
\endbibitem

\bibitem[\protect\citeauthoryear{Raballo et~al.}{2025}]{raballo2025diagnosing}
\begin{barticle}
\bauthor{\bsnm{Raballo}, \binits{A.}},
\bauthor{\bsnm{Ravenda}, \binits{F.}},
\bauthor{\bsnm{Mira}, \binits{A.}}:
\batitle{Diagnosing schizophrenia spectrum disorders: Large language models (llms) vs. leading international psychiatrists (lips)}.
\bjtitle{Psychiatry and Clinical Neurosciences}
\bvolume{79}(\bissue{9}),
\bfpage{599}
(\byear{2025})
\end{barticle}
\endbibitem

\bibitem[\protect\citeauthoryear{Tu et~al.}{2025}]{tu2025towards}
\begin{barticle}
\bauthor{\bsnm{Tu}, \binits{T.}},
\bauthor{\bsnm{Schaekermann}, \binits{M.}},
\bauthor{\bsnm{Palepu}, \binits{A.}},
\bauthor{\bsnm{Saab}, \binits{K.}},
\bauthor{\bsnm{Freyberg}, \binits{J.}},
\bauthor{\bsnm{Tanno}, \binits{R.}},
\bauthor{\bsnm{Wang}, \binits{A.}},
\bauthor{\bsnm{Li}, \binits{B.}},
\bauthor{\bsnm{Amin}, \binits{M.}},
\bauthor{\bsnm{Cheng}, \binits{Y.}}, \betal:
\batitle{Towards conversational diagnostic artificial intelligence}.
\bjtitle{Nature}
\bvolume{642}(\bissue{8067}),
\bfpage{442}--\blpage{450}
(\byear{2025})
\end{barticle}
\endbibitem

\bibitem[\protect\citeauthoryear{Li et~al.}{2024}]{li2024mediq}
\begin{barticle}
\bauthor{\bsnm{Li}, \binits{S.S.}},
\bauthor{\bsnm{Balachandran}, \binits{V.}},
\bauthor{\bsnm{Feng}, \binits{S.}},
\bauthor{\bsnm{Ilgen}, \binits{J.S.}},
\bauthor{\bsnm{Pierson}, \binits{E.}},
\bauthor{\bsnm{Koh}, \binits{P.W.}},
\bauthor{\bsnm{Tsvetkov}, \binits{Y.}}:
\batitle{Mediq: Question-asking llms and a benchmark for reliable interactive clinical reasoning}.
\bjtitle{Advances in Neural Information Processing Systems}
\bvolume{37},
\bfpage{28858}--\blpage{28888}
(\byear{2024})
\end{barticle}
\endbibitem

\bibitem[\protect\citeauthoryear{Katz et~al.}{2024}]{katz2024gpt}
\begin{barticle}
\bauthor{\bsnm{Katz}, \binits{D.M.}},
\bauthor{\bsnm{Bommarito}, \binits{M.J.}},
\bauthor{\bsnm{Gao}, \binits{S.}},
\bauthor{\bsnm{Arredondo}, \binits{P.}}:
\batitle{Gpt-4 passes the bar exam}.
\bjtitle{Philosophical transactions. Series A, Mathematical, physical, and engineering sciences}
\bvolume{382}(\bissue{2270}),
\bfpage{20230254}
(\byear{2024})
\end{barticle}
\endbibitem

\bibitem[\protect\citeauthoryear{Guha et~al.}{2023}]{guha2023legalbench}
\begin{barticle}
\bauthor{\bsnm{Guha}, \binits{N.}},
\bauthor{\bsnm{Nyarko}, \binits{J.}},
\bauthor{\bsnm{Ho}, \binits{D.}},
\bauthor{\bsnm{R{\'e}}, \binits{C.}},
\bauthor{\bsnm{Chilton}, \binits{A.}},
\bauthor{\bsnm{Chohlas-Wood}, \binits{A.}},
\bauthor{\bsnm{Peters}, \binits{A.}},
\bauthor{\bsnm{Waldon}, \binits{B.}},
\bauthor{\bsnm{Rockmore}, \binits{D.}},
\bauthor{\bsnm{Zambrano}, \binits{D.}}, \betal:
\batitle{Legalbench: A collaboratively built benchmark for measuring legal reasoning in large language models}.
\bjtitle{Advances in neural information processing systems}
\bvolume{36},
\bfpage{44123}--\blpage{44279}
(\byear{2023})
\end{barticle}
\endbibitem

\bibitem[\protect\citeauthoryear{Wu et~al.}{2023}]{wu2023bloomberggpt}
\begin{botherref}
\oauthor{\bsnm{Wu}, \binits{S.}},
\oauthor{\bsnm{Irsoy}, \binits{O.}},
\oauthor{\bsnm{Lu}, \binits{S.}},
\oauthor{\bsnm{Dabravolski}, \binits{V.}},
\oauthor{\bsnm{Dredze}, \binits{M.}},
\oauthor{\bsnm{Gehrmann}, \binits{S.}},
\oauthor{\bsnm{Kambadur}, \binits{P.}},
\oauthor{\bsnm{Rosenberg}, \binits{D.}},
\oauthor{\bsnm{Mann}, \binits{G.}}:
Bloomberggpt: A large language model for finance.
arXiv preprint arXiv:2303.17564
(2023)
\end{botherref}
\endbibitem

\bibitem[\protect\citeauthoryear{Chen et~al.}{2021}]{chen2021evaluating}
\begin{botherref}
\oauthor{\bsnm{Chen}, \binits{M.}},
\oauthor{\bsnm{Tworek}, \binits{J.}},
\oauthor{\bsnm{Jun}, \binits{H.}},
\oauthor{\bsnm{Yuan}, \binits{Q.}},
\oauthor{\bsnm{Pinto}, \binits{H.P.D.O.}},
\oauthor{\bsnm{Kaplan}, \binits{J.}},
\oauthor{\bsnm{Edwards}, \binits{H.}},
\oauthor{\bsnm{Burda}, \binits{Y.}},
\oauthor{\bsnm{Joseph}, \binits{N.}},
\oauthor{\bsnm{Brockman}, \binits{G.}}, et al.:
Evaluating large language models trained on code.
arXiv preprint arXiv:2107.03374
(2021)
\end{botherref}
\endbibitem

\bibitem[\protect\citeauthoryear{Kim et~al.}{2026}]{kim2026knowledge}
\begin{bchapter}
\bauthor{\bsnm{Kim}, \binits{J.}},
\bauthor{\bsnm{Pasini}, \binits{S.}},
\bauthor{\bsnm{Tonella}, \binits{P.}}:
\bctitle{When knowledge changes: Metamorphic testing of rag systems with mutations}.
In: \bbtitle{Proceedings of the 41st IEEE/ACM International Conference on Automated Software Engineering},
pp. \bfpage{3140}--\blpage{3151}
(\byear{2026})
\end{bchapter}
\endbibitem

\bibitem[\protect\citeauthoryear{Kasneci et~al.}{2023}]{kasneci2023chatgpt}
\begin{barticle}
\bauthor{\bsnm{Kasneci}, \binits{E.}},
\bauthor{\bsnm{Se{\ss}ler}, \binits{K.}},
\bauthor{\bsnm{K{\"u}chemann}, \binits{S.}},
\bauthor{\bsnm{Bannert}, \binits{M.}},
\bauthor{\bsnm{Dementieva}, \binits{D.}},
\bauthor{\bsnm{Fischer}, \binits{F.}},
\bauthor{\bsnm{Gasser}, \binits{U.}},
\bauthor{\bsnm{Groh}, \binits{G.}},
\bauthor{\bsnm{G{\"u}nnemann}, \binits{S.}},
\bauthor{\bsnm{H{\"u}llermeier}, \binits{E.}}, \betal:
\batitle{Chatgpt for good? on opportunities and challenges of large language models for education}.
\bjtitle{Learning and individual differences}
\bvolume{103},
\bfpage{102274}
(\byear{2023})
\end{barticle}
\endbibitem

\bibitem[\protect\citeauthoryear{Romera-Paredes et~al.}{2024}]{romera2024mathematical}
\begin{barticle}
\bauthor{\bsnm{Romera-Paredes}, \binits{B.}},
\bauthor{\bsnm{Barekatain}, \binits{M.}},
\bauthor{\bsnm{Novikov}, \binits{A.}},
\bauthor{\bsnm{Balog}, \binits{M.}},
\bauthor{\bsnm{Kumar}, \binits{M.P.}},
\bauthor{\bsnm{Dupont}, \binits{E.}},
\bauthor{\bsnm{Ruiz}, \binits{F.J.}},
\bauthor{\bsnm{Ellenberg}, \binits{J.S.}},
\bauthor{\bsnm{Wang}, \binits{P.}},
\bauthor{\bsnm{Fawzi}, \binits{O.}}, \betal:
\batitle{Mathematical discoveries from program search with large language models}.
\bjtitle{Nature}
\bvolume{625}(\bissue{7995}),
\bfpage{468}--\blpage{475}
(\byear{2024})
\end{barticle}
\endbibitem

\bibitem[\protect\citeauthoryear{Agrawal et~al.}{2022}]{agrawal2022large}
\begin{bchapter}
\bauthor{\bsnm{Agrawal}, \binits{M.}},
\bauthor{\bsnm{Hegselmann}, \binits{S.}},
\bauthor{\bsnm{Lang}, \binits{H.}},
\bauthor{\bsnm{Kim}, \binits{Y.}},
\bauthor{\bsnm{Sontag}, \binits{D.}}:
\bctitle{Large language models are few-shot clinical information extractors}.
In: \bbtitle{Proceedings of the 2022 Conference on Empirical Methods in Natural Language Processing},
pp. \bfpage{1998}--\blpage{2022}
(\byear{2022})
\end{bchapter}
\endbibitem

\bibitem[\protect\citeauthoryear{Van~Veen et~al.}{2024}]{van2024adapted}
\begin{barticle}
\bauthor{\bsnm{Van~Veen}, \binits{D.}},
\bauthor{\bsnm{Van~Uden}, \binits{C.}},
\bauthor{\bsnm{Blankemeier}, \binits{L.}},
\bauthor{\bsnm{Delbrouck}, \binits{J.-B.}},
\bauthor{\bsnm{Aali}, \binits{A.}},
\bauthor{\bsnm{Bluethgen}, \binits{C.}},
\bauthor{\bsnm{Pareek}, \binits{A.}},
\bauthor{\bsnm{Polacin}, \binits{M.}},
\bauthor{\bsnm{Reis}, \binits{E.P.}},
\bauthor{\bsnm{Seehofnerov{\'a}}, \binits{A.}}, \betal:
\batitle{Adapted large language models can outperform medical experts in clinical text summarization}.
\bjtitle{Nature medicine}
\bvolume{30}(\bissue{4}),
\bfpage{1134}--\blpage{1142}
(\byear{2024})
\end{barticle}
\endbibitem

\bibitem[\protect\citeauthoryear{McDuff et~al.}{2025}]{mcduff2025towards}
\begin{barticle}
\bauthor{\bsnm{McDuff}, \binits{D.}},
\bauthor{\bsnm{Schaekermann}, \binits{M.}},
\bauthor{\bsnm{Tu}, \binits{T.}},
\bauthor{\bsnm{Palepu}, \binits{A.}},
\bauthor{\bsnm{Wang}, \binits{A.}},
\bauthor{\bsnm{Garrison}, \binits{J.}},
\bauthor{\bsnm{Singhal}, \binits{K.}},
\bauthor{\bsnm{Sharma}, \binits{Y.}},
\bauthor{\bsnm{Azizi}, \binits{S.}},
\bauthor{\bsnm{Kulkarni}, \binits{K.}}, \betal:
\batitle{Towards accurate differential diagnosis with large language models}.
\bjtitle{Nature}
\bvolume{642}(\bissue{8067}),
\bfpage{451}--\blpage{457}
(\byear{2025})
\end{barticle}
\endbibitem

\bibitem[\protect\citeauthoryear{Goh et~al.}{2024}]{goh2024large}
\begin{barticle}
\bauthor{\bsnm{Goh}, \binits{E.}},
\bauthor{\bsnm{Gallo}, \binits{R.}},
\bauthor{\bsnm{Hom}, \binits{J.}},
\bauthor{\bsnm{Strong}, \binits{E.}},
\bauthor{\bsnm{Weng}, \binits{Y.}},
\bauthor{\bsnm{Kerman}, \binits{H.}},
\bauthor{\bsnm{Cool}, \binits{J.A.}},
\bauthor{\bsnm{Kanjee}, \binits{Z.}},
\bauthor{\bsnm{Parsons}, \binits{A.S.}},
\bauthor{\bsnm{Ahuja}, \binits{N.}}, \betal:
\batitle{Large language model influence on diagnostic reasoning: a randomized clinical trial}.
\bjtitle{JAMA network open}
\bvolume{7}(\bissue{10}),
\bfpage{2440969}
(\byear{2024})
\end{barticle}
\endbibitem

\bibitem[\protect\citeauthoryear{Brodeur et~al.}{2024}]{brodeur2024superhuman}
\begin{botherref}
\oauthor{\bsnm{Brodeur}, \binits{P.}},
\oauthor{\bsnm{Buckley}, \binits{T.}},
\oauthor{\bsnm{Kanjee}, \binits{Z.}},
\oauthor{\bsnm{Goh}, \binits{E.}},
\oauthor{\bsnm{Ling}, \binits{E.}},
\oauthor{\bsnm{Jain}, \binits{P.}},
\oauthor{\bsnm{Cabral}, \binits{S.}},
\oauthor{\bsnm{Abdulnour}, \binits{R.}},
\oauthor{\bsnm{Haimovich}, \binits{A.}},
\oauthor{\bsnm{Freed}, \binits{J.}}, et al.:
Superhuman performance of a large language model on the reasoning tasks of a physician. arxiv 2024.
arXiv preprint arXiv:2412.10849
(2024)
\end{botherref}
\endbibitem

\bibitem[\protect\citeauthoryear{Vrdoljak et~al.}{2025}]{vrdoljak2025evaluating}
\begin{barticle}
\bauthor{\bsnm{Vrdoljak}, \binits{J.}},
\bauthor{\bsnm{Boban}, \binits{Z.}},
\bauthor{\bsnm{Males}, \binits{I.}},
\bauthor{\bsnm{Skrabic}, \binits{R.}},
\bauthor{\bsnm{Kumric}, \binits{M.}},
\bauthor{\bsnm{Ottosen}, \binits{A.}},
\bauthor{\bsnm{Clemencau}, \binits{A.}},
\bauthor{\bsnm{Bozic}, \binits{J.}},
\bauthor{\bsnm{V{\"o}lker}, \binits{S.}}:
\batitle{Evaluating large language and large reasoning models as decision support tools in emergency internal medicine}.
\bjtitle{Computers in biology and medicine}
\bvolume{192},
\bfpage{110351}
(\byear{2025})
\end{barticle}
\endbibitem

\bibitem[\protect\citeauthoryear{Bhayana}{2024}]{bhayana2024chatbots}
\begin{barticle}
\bauthor{\bsnm{Bhayana}, \binits{R.}}:
\batitle{Chatbots and large language models in radiology: a practical primer for clinical and research applications}.
\bjtitle{Radiology}
\bvolume{310}(\bissue{1}),
\bfpage{232756}
(\byear{2024})
\end{barticle}
\endbibitem

\bibitem[\protect\citeauthoryear{Goktas and Grzybowski}{2024}]{goktas2024assessing}
\begin{barticle}
\bauthor{\bsnm{Goktas}, \binits{P.}},
\bauthor{\bsnm{Grzybowski}, \binits{A.}}:
\batitle{Assessing the impact of chatgpt in dermatology: a comprehensive rapid review}.
\bjtitle{Journal of Clinical Medicine}
\bvolume{13}(\bissue{19}),
\bfpage{5909}
(\byear{2024})
\end{barticle}
\endbibitem

\bibitem[\protect\citeauthoryear{Bedi et~al.}{2026}]{bedi2026holistic}
\begin{barticle}
\bauthor{\bsnm{Bedi}, \binits{S.}},
\bauthor{\bsnm{Cui}, \binits{H.}},
\bauthor{\bsnm{Fuentes}, \binits{M.}},
\bauthor{\bsnm{Unell}, \binits{A.}},
\bauthor{\bsnm{Wornow}, \binits{M.}},
\bauthor{\bsnm{Banda}, \binits{J.M.}},
\bauthor{\bsnm{Kotecha}, \binits{N.}},
\bauthor{\bsnm{Keyes}, \binits{T.}},
\bauthor{\bsnm{Mai}, \binits{Y.}},
\bauthor{\bsnm{Oez}, \binits{M.}}, \betal:
\batitle{Holistic evaluation of large language models for medical tasks with medhelm}.
\bjtitle{Nature medicine}
\bvolume{32}(\bissue{3}),
\bfpage{943}--\blpage{951}
(\byear{2026})
\end{barticle}
\endbibitem

\bibitem[\protect\citeauthoryear{Arora et~al.}{2025}]{arora2025healthbench}
\begin{botherref}
\oauthor{\bsnm{Arora}, \binits{R.K.}},
\oauthor{\bsnm{Wei}, \binits{J.}},
\oauthor{\bsnm{Hicks}, \binits{R.S.}},
\oauthor{\bsnm{Bowman}, \binits{P.}},
\oauthor{\bsnm{Qui{\~n}onero-Candela}, \binits{J.}},
\oauthor{\bsnm{Tsimpourlas}, \binits{F.}},
\oauthor{\bsnm{Sharman}, \binits{M.}},
\oauthor{\bsnm{Shah}, \binits{M.}},
\oauthor{\bsnm{Vallone}, \binits{A.}},
\oauthor{\bsnm{Beutel}, \binits{A.}}, et al.:
Healthbench: Evaluating large language models towards improved human health.
arXiv preprint arXiv:2505.08775
(2025)
\end{botherref}
\endbibitem

\bibitem[\protect\citeauthoryear{Wu et~al.}{2026}]{wu2026medarena}
\begin{botherref}
\oauthor{\bsnm{Wu}, \binits{E.}},
\oauthor{\bsnm{Wu}, \binits{K.}},
\oauthor{\bsnm{Hom}, \binits{J.}},
\oauthor{\bsnm{Yi}, \binits{P.H.}},
\oauthor{\bsnm{Zhang}, \binits{A.}},
\oauthor{\bsnm{Lozano}, \binits{A.}},
\oauthor{\bsnm{Nirschl}, \binits{J.}},
\oauthor{\bsnm{Tangney}, \binits{J.}},
\oauthor{\bsnm{Byram}, \binits{K.}},
\oauthor{\bsnm{Dymm}, \binits{B.}}, et al.:
Medarena: Comparing llms for medicine-in-the-wild clinician preferences.
arXiv preprint arXiv:2603.15677
(2026)
\end{botherref}
\endbibitem

\bibitem[\protect\citeauthoryear{Elhassan et~al.}{2026}]{elhassan2026preferred}
\begin{botherref}
\oauthor{\bsnm{Elhassan}, \binits{F.}},
\oauthor{\bsnm{Sasu}, \binits{D.}},
\oauthor{\bsnm{Kulinkina}, \binits{A.}},
\oauthor{\bsnm{Klein}, \binits{L.H.}},
\oauthor{\bsnm{Hartley}, \binits{M.-A.}}:
Preferred, not safer: Pairwise preference is a poor proxy for clinical safety.
arXiv preprint arXiv:2608.02617
(2026)
\end{botherref}
\endbibitem

\bibitem[\protect\citeauthoryear{Schmidgall et~al.}{2026}]{schmidgall2026agentclinic}
\begin{botherref}
\oauthor{\bsnm{Schmidgall}, \binits{S.}},
\oauthor{\bsnm{Ziaei}, \binits{R.}},
\oauthor{\bsnm{Harris}, \binits{C.}},
\oauthor{\bsnm{Kim}, \binits{J.W.}},
\oauthor{\bsnm{Reis}, \binits{E.P.}},
\oauthor{\bsnm{Jopling}, \binits{J.}},
\oauthor{\bsnm{Moor}, \binits{M.}}:
Agentclinic: a multimodal benchmark for tool-using clinical ai agents.
npj Digital Medicine
(2026)
\end{botherref}
\endbibitem

\bibitem[\protect\citeauthoryear{Johri et~al.}{2025}]{johri2025evaluation}
\begin{barticle}
\bauthor{\bsnm{Johri}, \binits{S.}},
\bauthor{\bsnm{Jeong}, \binits{J.}},
\bauthor{\bsnm{Tran}, \binits{B.A.}},
\bauthor{\bsnm{Schlessinger}, \binits{D.I.}},
\bauthor{\bsnm{Wongvibulsin}, \binits{S.}},
\bauthor{\bsnm{Barnes}, \binits{L.A.}},
\bauthor{\bsnm{Zhou}, \binits{H.-Y.}},
\bauthor{\bsnm{Cai}, \binits{Z.R.}},
\bauthor{\bsnm{Van~Allen}, \binits{E.M.}},
\bauthor{\bsnm{Kim}, \binits{D.}}, \betal:
\batitle{An evaluation framework for clinical use of large language models in patient interaction tasks}.
\bjtitle{Nature medicine}
\bvolume{31}(\bissue{1}),
\bfpage{77}--\blpage{86}
(\byear{2025})
\end{barticle}
\endbibitem

\bibitem[\protect\citeauthoryear{Niu et~al.}{2026}]{niu2026aipatient}
\begin{botherref}
\oauthor{\bsnm{Niu}, \binits{J.}},
\oauthor{\bsnm{Yu}, \binits{H.}},
\oauthor{\bsnm{Wang}, \binits{W.}},
\oauthor{\bsnm{Dai}, \binits{G.}},
\oauthor{\bsnm{He}, \binits{J.}},
\oauthor{\bsnm{Li}, \binits{X.}},
\oauthor{\bsnm{Liang}, \binits{Z.}},
\oauthor{\bsnm{Lin}, \binits{X.}},
\oauthor{\bsnm{So}, \binits{K.C.}},
\oauthor{\bsnm{Yan}, \binits{B.Y.}}, et al.:
Aipatient arena: Ehr-grounded evaluation of large language models in end-to-end clinical consultation workflows.
arXiv preprint arXiv:2606.17474
(2026)
\end{botherref}
\endbibitem

\bibitem[\protect\citeauthoryear{Chen et~al.}{2023}]{chen2023put}
\begin{botherref}
\oauthor{\bsnm{Chen}, \binits{J.}},
\oauthor{\bsnm{Yuan}, \binits{S.}},
\oauthor{\bsnm{Ye}, \binits{R.}},
\oauthor{\bsnm{Majumder}, \binits{B.P.}},
\oauthor{\bsnm{Richardson}, \binits{K.}}:
Put your money where your mouth is: Evaluating strategic planning and execution of llm agents in an auction arena.
arXiv preprint arXiv:2310.05746
(2023)
\end{botherref}
\endbibitem

\bibitem[\protect\citeauthoryear{Hampton et~al.}{1975}]{hampton1975relative}
\begin{barticle}
\bauthor{\bsnm{Hampton}, \binits{J.R.}},
\bauthor{\bsnm{Harrison}, \binits{M.}},
\bauthor{\bsnm{Mitchell}, \binits{J.R.}},
\bauthor{\bsnm{Prichard}, \binits{J.S.}},
\bauthor{\bsnm{Seymour}, \binits{C.}}:
\batitle{Relative contributions of history-taking, physical examination, and laboratory investigation to diagnosis and management of medical outpatients.}
\bjtitle{British medical journal}
\bvolume{2}(\bissue{5969}),
\bfpage{486}--\blpage{489}
(\byear{1975})
\end{barticle}
\endbibitem

\end{thebibliography}

\end{document}